\documentclass[sigconf]{acmart}
\usepackage{balance}
\def\BibTeX{{\rm B\kern-.05em{\sc i\kern-.025em b}\kern-.08emT\kern-.1667em\lower.7ex\hbox{E}\kern-.125emX}}

\usepackage{microtype}
\usepackage{graphicx}
\usepackage{subfigure}
\usepackage{booktabs} % for professional tables
\usepackage{amsmath}
\usepackage{amssymb}
\usepackage{graphicx}
\usepackage[ruled,vlined, linesnumbered]{algorithm2e}

\usepackage{enumitem}
\setlist[enumerate]{leftmargin=11pt}
\setlist[itemize]{leftmargin=11pt}

\usepackage{hyperref}

\def\Y{\mathcal{Y}}
\def\B{\mathcal{B}}

\copyrightyear{2019} 
\acmYear{2019} 
\setcopyright{rightsretained} 
\acmConference[KDD '19]{The 25th ACM SIGKDD Conference on Knowledge Discovery and Data Mining}{August 4--8, 2019}{Anchorage, AK, USA}
\acmBooktitle{The 25th ACM SIGKDD Conference on Knowledge Discovery and Data Mining (KDD '19), August 4--8, 2019, Anchorage, AK, USA}
\acmDOI{10.1145/3292500.3330925}
\acmISBN{978-1-4503-6201-6/19/08}
\begin{document}

%
% The "title" command has an optional parameter, allowing the author to define a "short title" to be used in page headers.
\title{Cluster-GCN: An Efficient Algorithm for Training Deep and Large Graph Convolutional Networks}

%
% The "author" command and its associated commands are used to define the authors and their affiliations.
% Of note is the shared affiliation of the first two authors, and the "authornote" and "authornotemark" commands
% used to denote shared contribution to the research.
% \author{Anonymous Author(s)}
% \affiliation{%

% \affiliation{%

%
% By default, the full list of authors will be used in the page headers. Often, this list is too long, and will overlap
% other information printed in the page headers. This command allows the author to define a more concise list
% of authors' names for this purpose.

\author{Wei-Lin Chiang}
\authornote{This work was done during the first and the second author's internship at Google Research.}
\affiliation{
  \institution{National Taiwan University}
}
\email{r06922166@csie.ntu.edu.tw}

\author{Xuanqing Liu}
\authornotemark[1]
\affiliation{
  \institution{University of California, Los Angeles}
}
\email{xqliu@cs.ucla.edu}

\author{Si Si}
\affiliation{
  \institution{Google Research}
}
\email{sisidaisy@google.com}

\author{Yang Li}
\affiliation{
  \institution{Google Research}
}
\email{liyang@google.com}
 
\author{Samy Bengio}
\affiliation{
  \institution{Google Research}
}
\email{bengio@google.com}

\author{Cho-Jui Hsieh}
\affiliation{
  \institution{University of California, Los Angeles}
}
\email{chohsieh@cs.ucla.edu}

\renewcommand{\shortauthors}{Chiang, et al.}
%
% The abstract is a short summary of the work to be presented in the article.
\begin{abstract}
Graph convolutional network (GCN) has been successfully applied to many graph-based applications; however, training a large-scale GCN remains challenging. Current SGD-based algorithms suffer from either a high computational cost that exponentially grows with  number of GCN layers, or a large space requirement for keeping the entire graph and the embedding of each node in memory. 
%the embeddings of all nodes in memory. 
%The main bottleneck in current GCN training is the exponential grow of neighborhood sampling, which makes previous algorithms having computational complexity exponential to number of layers. 
In this paper, we propose Cluster-GCN, a novel GCN algorithm that is suitable for SGD-based training by exploiting the graph clustering structure. 
%Motivating by the observation that the efficiency of mini-batch GCN training is controlled by number of within-batch links, our algorithm
%partition the graph using graph clustering algorithms and then 
%We found the efficiency of GCN mini-batch training is controlled by numer of within-batch links, which motivates the use of resolve this issue by exploiting the clustering structure of graphs.
Cluster-GCN works as the following: at each step, it samples a block of nodes that associate with a dense subgraph identified by a graph clustering algorithm, and restricts the neighborhood search within this subgraph. This simple but effective strategy leads to significantly improved memory and computational efficiency while being able to achieve comparable test accuracy with previous algorithms. 
To test the scalability of our algorithm, we create a new Amazon2M data with 2 million nodes and 61 million edges which is more than 5 times larger than the previous largest publicly available dataset (Reddit). For training a 3-layer GCN on this data, Cluster-GCN is faster than the previous state-of-the-art VR-GCN (1523 seconds vs 1961 seconds) and using much less memory (2.2GB vs 11.2GB). Furthermore, for training 4 layer GCN on this data, our algorithm  can finish in around 36 minutes while all the existing GCN training algorithms fail to train due to the out-of-memory issue. Furthermore, Cluster-GCN allows us to train much deeper GCN without much time and memory overhead, which leads to improved prediction accuracy---using a 5-layer Cluster-GCN, we achieve state-of-the-art test F1 score 99.36 on the PPI dataset, while the previous best result was 98.71 by~\cite{zhang2018gaan}.
Our codes are publicly available at \url{https://github.com/google-research/google-research/tree/master/cluster_gcn}.
%As an example, on Amazon2M data, which contains a product co-purchase graph with more than 2 million nodes and 61 millions edges, a 3-layer Cluster-GCN can achieve test F1 score 90.21, which is the same accurate as VRGCN(\cite{chen2018stochastic}) while using less training time (1523 seconds vs 1961 seconds) and much less memory (2.2GB vs 11.2GB). Furthermore,  a 4-layer Cluster-GCN can even boost the F1 score to 90.41 using around 36 minutes and 2G memory for training, while all the existing GCN training algorithms fail to train due to the OOM issue. 
%Furthermore, our method has time complexity linear to number of layers without any memory overhead, while previous approaches either have complexity exponential to number of layers or require huge amount of additional memory for caching temporary variables during training. Experimental results on large-scale graph data with millions of nodes demonstrate the scalability of our method.
\end{abstract}



%
% Keywords. The author(s) should pick words that accurately describe the work being
% presented. Separate the keywords with commas.
% \keywords{datasets, neural networks, gaze detection, text tagging}

\maketitle

\section{Introduction}
\label{sec:intro}

Graph convolutional network (GCN)~\cite{kipf2017semi} has become increasingly popular in addressing many graph-based applications, including semi-supervised node classification~\cite{kipf2017semi}, link prediction~\cite{zhang2018link} and recommender systems~\cite{Ying:2018}. 
Given a graph, GCN uses a graph convolution operation to obtain node embeddings layer by layer---at each layer, the embedding of a node is obtained by gathering the embeddings of its neighbors, followed by one or a few layers of linear transformations and nonlinear activations. The final layer embedding is then used for some end tasks. For instance, in node classification problems, the final layer embedding is passed to a classifier to predict node labels, and thus the parameters of GCN can be trained in an end-to-end manner. 
%f GCN can be trained end-to-end Some supervised information, for instance, labels observed at a subset of nodes, are given to form the loss function based on the final layer node embeddings. 

Since the graph convolution operator in GCN needs to propagate embeddings using the interaction between nodes in the graph, this makes training quite challenging. Unlike other neural networks that the training loss can be perfectly decomposed into individual terms on each sample, the loss term in GCN (e.g., classification loss on a single node)  depends on a huge number of other nodes, especially when GCN goes deep. Due to the node dependence, GCN's training is very slow and requires lots of memory -- back-propagation needs to store all the embeddings in the computation graph in GPU memory.
%, and when training in GPU it is time consuming to swap between GPU and CPU memory. 
%many GCN training algorithms have huge memory requirement, which has become a bottleneck for scaling especially in GPU-based training. 

{\bf Previous GCN Training Algorithms:  } 
To demonstrate the need of developing a scalable GCN training algorithm, we first discuss the pros and cons of existing approaches, in terms of 1) memory requirement{\footnote{Here we consider the memory for storing node embeddings, which is dense and usually dominates the overall memory usage for deep GCN.}}, 2) time per epoch\footnote{An epoch means a complete data pass. } and 3) convergence speed  (loss reduction) per epoch. These three factors are crucial for evaluating a training algorithm. Note that memory requirement directly restricts the scalability of algorithm, and the later two factors combined together will determine the training speed. In the following discussion we denote $N$ to be the number of nodes in the graph, $F$ the embedding dimension, and $L$ the number of layers to analyze classic GCN training algorithms. 
\begin{itemize}
\item Full-batch gradient descent is proposed in the first GCN paper~\cite{kipf2017semi}. To compute the full gradient, it requires storing all the intermediate embeddings, leading to $O(NFL)$ memory requirement, which is not scalable. Furthermore, although the time per epoch is efficient, the convergence of gradient descent is slow since the parameters are updated only once per epoch.\\
$[$memory: bad; time per epoch: good; convergence: bad$]$
\item Mini-batch SGD is proposed in~\cite{hamilton2017inductive}. Since each update is only based on a mini-batch gradient, it can reduce the memory requirement and conduct many updates per epoch, leading to a faster convergence. However, mini-batch SGD introduces a significant computational overhead due to the {\bf neighborhood expansion problem}---to compute the loss on a single node at layer $L$, it requires that node's neighbor nodes' embeddings at layer $L-1$, which again requires their neighbors' embeddings at layer $L-2$ and recursive ones in the downstream layers. This leads to time complexity exponential to the GCN depth. GraphSAGE~\cite{hamilton2017inductive} proposed to use a fixed size of neighborhood samples during back-propagation through layers and FastGCN~\cite{chen2018fastgcn}  proposed importance sampling, but the overhead of these methods is still large and will become worse when GCN goes deep.\\
$[$memory: good; time per epoch: bad; convergence: good$]$
 \item VR-GCN~\cite{chen2018stochastic} proposes to use a variance reduction technique to reduce the size of neighborhood sampling nodes. Despite successfully reducing the size of samplings (in our experiments VR-GCN with only 2 samples per node works quite well), it requires storing all the intermediate embeddings of all the nodes in memory, leading to $O(NFL)$ memory requirement.
 If the number of nodes in the graph increases to millions, the memory requirement for VR-GCN may be too high to fit into GPU. 
 \\
 {$[$memory: bad; time per epoch: good; convergence: good.$]$}
\end{itemize}

In this paper, we propose a novel GCN training algorithm by exploiting the graph clustering structure. We find that the efficiency of a mini-batch algorithm can be characterized by the notion of ``embedding utilization'', which is proportional to the number of links between nodes in one batch or within-batch links. This finding motivates us to design the batches using graph clustering algorithms that aims to construct partitions of nodes so that there are more graph links between nodes in the same partition than nodes in different partitions. Based on the graph clustering idea, we proposed Cluster-GCN, an algorithm to design the batches based on efficient graph clustering algorithms (e.g., METIS~\citep{Metis}). We take this idea further by proposing a stochastic multi-clustering framework to improve the convergence of Cluster-GCN. Our strategy leads to huge memory and computational benefits. In terms of memory, we only need to store the node embeddings within the current batch, which is $O(bFL)$ with the batch size $b$. This is significantly better than VR-GCN and full gradient decent, and slightly better than other SGD-based approaches. In terms of computational complexity, our algorithm achieves the same time cost per epoch with gradient descent and is much faster than neighborhood searching approaches. In terms of the convergence speed, our algorithm is competitive with other SGD-based approaches. Finally, our algorithm is simple to implement since we only compute matrix multiplication and no neighborhood sampling is needed. Therefore for Cluster-GCN, we have
$[$memory: good; time per epoch: good; convergence: good$]$.

We conducted comprehensive experiments on several large-scale graph datasets and made the following contributions: 
\begin{itemize}
\item Cluster-GCN achieves the best memory usage on large-scale graphs, especially on deep GCN. For example, Cluster-GCN uses 5x less memory than VRGCN in a 3-layer GCN model on Amazon2M. Amazon2M is a new graph dataset that we construct to demonstrate the scalablity of the GCN algorithms. This dataset contains a amazon product co-purchase graph with more than 2 millions nodes and 61 millions edges.
\item Cluster-GCN achieves a similar training speed with VR-GCN for shallow networks (e.g., 2 layers) but can be faster than VR-GCN when the network goes deeper (e.g., 4 layers), since our complexity is linear to the number of layers $L$ while VR-GCN's complexity is exponential to $L$.
\item Cluster-GCN is able to train a very deep network that has a large embedding size. Although several previous works show that deep GCN does not give better performance, we found that with proper optimization, deeper GCN could help the accuracy. For example, with a 5-layer GCN, we obtain a new benchmark accuracy 99.36 for PPI dataset, comparing with the highest reported one 98.71 by \cite{zhang2018gaan}.
\end{itemize}
Implementation of our proposed method is publicly available.\footnote{\url{https://github.com/google-research/google-research/tree/master/cluster\_gcn}}
%The rest of the paper is organized as follows. In Section 2, we introduce GCN and discuss the memory and computation bottlenecks. We then propose a cluster-GCN algorithm in Section 3 to resolve these issues and show the experimental results in Section 4. We conclude the paper in Section 5. 

\section{Background}
 Suppose we are given a graph $G =(\mathcal{V},\mathcal{E},A)$, which consists 
of $N=|\mathcal{V}|$ vertices and $|\mathcal{E}|$ edges such that an edge between any two 
vertices $i$ and $j$ represents their similarity. The corresponding 
adjacency matrix $A$ is an $N \times N$ sparse matrix 
with $(i,j)$ entry equaling to 1 if there is an edge between $i$ and $j$ 
and $0$ otherwise. Also, each node is associated with an $F$-dimensional feature vector and $X\in \mathbb{R}^{N \times F}$  denotes the feature matrix for all $N$ nodes.
An $L$-layer GCN~\cite{kipf2017semi} consists of $L$ graph convolution layers and each of them constructs embeddings for each node by mixing the embeddings of the node's neighbors in the graph from the previous layer:
\begin{equation}
    Z^{(l+1)} = A'X^{(l)}W^{(l)}, \ \ 
    X^{(l+1)} = \sigma(Z^{(l+1)}), 
    \label{eq:gcn0}
\end{equation}
where $X^{(l)}\in \mathbb{R}^{N \times F_l}$ is the embedding at the $l$-th layer for all the $N$ nodes and $X^{(0)}=X$;  %$X^{l+1}$ is the embedding for $l+1$-layer; 
$A'$ is the normalized and regularized adjacency matrix 
%$A' =
%\tilde{D}^{-1} \tilde{A}$
%\tilde{D}^{-\frac{1}{2}} \tilde{A} \tilde{D}^{-\frac{1}{2}}$ 
%and $\tilde{A} = A + I, \tilde{D}_{ii} = \sum_{j} \tilde{A}_{ij}$;  
and $W^{(l)}\in \mathbb{R}^{F_l \times F_{l+1}}$ is the feature transformation matrix which will be learnt for the downstream tasks.
Note that for simplicity we assume the feature dimensions are the same for all layers ($F_1=\dots = F_L = F$). 
The activation function $\sigma(\cdot)$ is usually set to be the element-wise ReLU. 

Semi-supervised node classification is a popular application of GCN. When using GCN for this application, the goal is to learn weight matrices in \eqref{eq:gcn0} by minimizing the loss function:
\begin{equation}
\mathcal{L} = \frac{1}{|\Y_L|} \sum_{i\in \Y_L} \text{loss}(y_i, z^{L}_i ), 
\label{eq:gcn-loss}
\end{equation}
where $\Y_L$ contains all the labels for the labeled nodes; $z^{(L)}_i$ is the $i$-th row of $Z^{(L)}$ with the ground-truth label to be $y_i$, indicating the final layer prediction of node $i$. In practice, a cross-entropy loss is commonly used for node classification in multi-class or multi-label problems.
%In this paper, we consider the GCN training problem, and the goal is to learn $\{W^{(l)}\}_{l=1}^L$ by minimizing the objective function \eqref{eq:gcn-loss}.
%(only one ``1'' for each $y_i$) or multi-label (more than one ``1''s for each $y_i$) problems. 
%which can be written as
%\begin{equation*}
%    \text{loss}(y, z) = \sum_{j} y_j \log \text{softmax}(z)_j. 
%\end{equation*}
%Given graph $A$, feature matrix $X$ and an additional label vector $Y \in \mathbb{R}^{N}$ for $N$ nodes and labels $Y_{L}$ are available, we can construct several layers of GCN to predict the label for 
%
%solve SSL problem. As a simple case, a 2-layer GCN\cite{kipf2017semi} will be
%\begin{equation}
%    Z = f(X, A) = \text{softmax} \left( \hat{A} \ \text{ReLU} \left(\hat{A} X W^{(0)}\right) \ W^{(1)} \right)
%\end{equation}
%
%followed by a cross-entropy loss over all labeled examples;
%\begin{equation} \label{eq:gcn-loss}
%    \mathcal{L} = -\sum_{l \in Y_{L}} \sum_{f=1}^{F} Y_{lf} \ln{Z_{lf}}
%\end{equation}

%Traditionally, GCN adopts a full-batch gradient descent to optimize \eqref{eq:gcn-loss}, which will bring scalability issues due to heavy neighbors finding which is exponentially growing w.r.t. the layers. Next, we will explain how to speed up GCN training by exploiting the graph structure.

\begin{table*}[t]
    \centering
        \caption{Time and space complexity of GCN training algorithms. $L$ is number of layers, $N$ is number of nodes, $\|A\|_0$ is number of nonzeros in the adjacency matrix, and $F$ is number of features. For simplicity we assume number of features is fixed for all layers. For SGD-based approaches, $b$ is the batch size and $r$ is the number of sampled neighbors per node. Note that due to the variance reduction technique, VR-GCN can work with a smaller $r$ than GraphSAGE and FastGCN.
        For memory complexity, $LF^2$ is for storing $\{W^{(l)}\}_{l=1}^L$ and the other term is for storing embeddings. For simplicity we omit the memory for storing the graph (GCN) or sub-graphs (other approaches) since they are fixed and usually not the main bottleneck.}
    \label{tab:complexity}
    \vspace{-10pt}
     \resizebox{0.8\textheight}{!}{
    \begin{tabular}{c|c|c|c|c|c|c}
         &  GCN~\cite{kipf2017semi} &
         Vanilla SGD
         &
         GraphSAGE~\cite{hamilton2017inductive} & FastGCN~\cite{chen2018fastgcn} &
         VR-GCN~\cite{chen2018stochastic} & Cluster-GCN  \\
         \hline 
         Time complexity & {\bf $O(L \|A\|_0 F + LNF^2 )$} &
         $O(d^L NF^2)$ & $O(r^LNF^2)$ & $O(rL NF^2)$ & $O(L \|A\|_0 F + LNF^2 + r^L NF^2 )$ &  {\bf $O(L\|A\|_0 F + LNF^2)$} \\
         Memory complexity &  $O(LNF+LF^2)$ & 
         $O(bd^LF+LF^2)$
         & $O(br^L F+LF^2)$ & $O(brLF+LF^2)$ & $O(LNF+LF^2)$ & {\bf $O(bLF+LF^2)$}
    \end{tabular}
    }

\end{table*}

\section{Proposed Algorithm}
\label{sec:algorithm}

We first discuss the bottleneck of previous training methods to motivate the proposed algorithm. 

%\subsubsection*{\underline{Why is full gradient descent slow for GCN training?}}
In the original paper~\cite{kipf2017semi}, full gradient descent is used for training GCN, but it suffers from high computational and memory cost. In terms of memory, computing the full gradient of \eqref{eq:gcn-loss} by back-propagation requires storing all the embedding matrices $\{Z^{(l)}\}_{l=1}^L$ which needs $O(NFL)$ space. In terms of convergence speed, since the model is only updated once per epoch, the training requires more epochs to converge.

%The original G
%In the original GCN paper~\cite{kipf2017semi}, 
%Traditional training strategy for GCN is to use full gradient, and suffers from scalability issues. 

It has been shown that mini-batch SGD can improve the training speed and memory requirement of GCN in some recent works~\citep{hamilton2017inductive, chen2018fastgcn, chen2018stochastic}. 
Instead of computing the full gradient, SGD only needs to calculate the gradient based on a mini-batch for each update.
In this paper, we use  $\B\subseteq [N]$ with size $b=|\B|$ to denote a batch of node indices, and each SGD step will compute the gradient estimation
\begin{equation}
 \frac{1}{|\B|} \sum_{i\in \B} \nabla \text{loss}(y_i, z^{(L)}_i ) 
\label{eq:sgd-loss}
\end{equation}
to perform an update. 
%, $B$ needs to aggregate each node $i$'s neighbors ${N_{i}}=[ {i_1},\cdots, {i_k}]$'s features and $k$ is the number of neighbors for node $i$. 
%Therefore, it can update the model multiple times per epoch.
Despite faster convergence in terms of epochs, SGD will introduce another computational overhead on GCN training (as explained in the following), which makes it having much slower per-epoch time compared with full gradient descent. 
\noindent\paragraph{\bf {Why does vanilla mini-batch SGD  have slow per-epoch time? }} 
We consider the computation of the gradient associated with one node $i: \nabla\text{loss}(y_i, z^{(L)}_i)$. 
Clearly, this requires  the embedding of node $i$, which depends on its neighbors' embeddings in the previous layer. 
%Since GCN has many layers, 
To fetch each node $i$'s neighbor nodes' embeddings, we need to further aggregate each neighbor node's %$\{i_j\}_{j=1}^{k}$'s 
neighbor nodes' embeddings as well.
Suppose a GCN has $L+1$ layers and each node has an average degree of $d$, to get the gradient for node $i$, we need to aggregate features from $O(d^L)$ nodes in the graph for one node. That is, we need to fetch information for a node's hop-$k$ ($k=1, \cdots, L$) neighbors in the graph to perform one update.
%Therefore, {\bf for computing the gradient of one node, we need to compute $O(d^L)$ embeddings in average}.  
Computing each embedding requires $O(F^2)$ time due to the multiplication with $W^{(l)}$, so in average computing the gradient associated with one node requires $O(d^LF^2)$ time.

{\bf Embedding utilization can reflect computational efficiency. }
If a batch has more than one node, the time complexity is less straightforward since different nodes can have overlapped hop-$k$ neighbors, and the number of embedding computation can be less than the worst case $O(b d^L)$. 
To reflect the computational efficiency of mini-batch SGD, we define the concept of {\bf ``embedding utilization''} to characterize the computational efficiency. During the algorithm, if the node $i$'s embedding at $l$-th layer $z^{(l)}_i$ is computed and is reused $u$ times for the embedding computations at layer $l+1$, then we say the embedding utilization of $z^{(l)}_i$ is $u$. For mini-batch SGD with random sampling, $u$ is very small since the graph is usually large and sparse. 
Assume $u$ is a small constant (almost no overlaps between hop-$k$ neighbors), then mini-batch SGD needs to compute $O(bd^L)$ embeddings per batch, which leads to $O(bd^LF^2)$ time per update and $O(Nd^LF^2)$ time per epoch. 

\begin{figure}
 \centering 
    \includegraphics[width=0.9\linewidth]{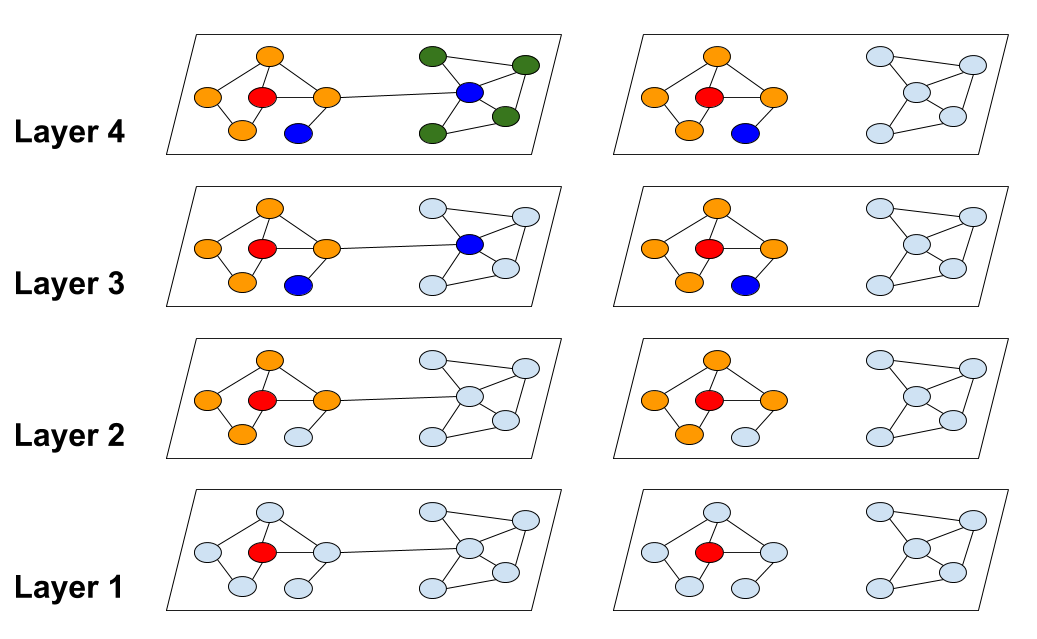}
    \caption{The neighborhood expansion difference between traditional
    graph convolution and our proposed cluster approach. The red node is the starting node for neighborhood nodes expansion. Traditional graph convolution suffers from exponential neighborhood expansion, while our method  can avoid expensive neighborhood expansion.}
    \label{fig:cluster}
    \vspace{-10pt}
\end{figure}

We illustrate the neighborhood expansion problem in the left panel of Fig.~\ref{fig:cluster}. 
In contrary, full-batch gradient descent has the maximal embedding utilization---each embedding will be reused $d$ (average degree) times in the upper layer. 
%When batch size $b$ is small, although the nodes in the current batch can sometimes have overlapped hop-$k$ neighbors, the overlapping is usually neglectable for a large-scale sparse graph. 
As a consequence, the original full gradient descent~\cite{kipf2017semi} only needs to compute $O(NL)$ embeddings per epoch, which means on average only $O(L)$ embedding computation is needed to acquire the gradient of one node.  
%The main reason is that in full gradient descent, each computed embedding in the intermediate layer is reused by multiple loss terms, while in SGD, due to the small minibatch, each computed embedding is only used by one or few loss terms, leading to {\bf low utilization of each node embedding computation}.  

%This is also known as the node expansion issue. 
%Furthermore, from the implementation perspective, each node's neighborhood node searching process across many layers is the most time-consuming part when the graph is large, and is hard to be accelerated by GPU  and slow to implement in python. 
%As a consequence, vanilla mini-batch training brings heavy overhead for each update, leading to slow GCN training. 

To make mini-batch SGD work, previous approaches try to restrict the neighborhood expansion size, which however do not improve embedding utilization.  GraphSAGE~\cite{hamilton2017inductive} 
 uniformly samples a fixed-size set of neighbors, instead of
using a full-neighborhood set. We denote the sample size as $r$. This leads to $O(r^L)$ embedding computations for each loss term but also makes gradient estimation less accurate. 
FastGCN~\cite{chen2018fastgcn} proposed an important sampling strategy to improve the gradient estimation. VR-GCN~\cite{chen2018stochastic} proposed a strategy to store the previous computed embeddings for all the $N$ nodes and $L$ layers and reuse them for unsampled neighbors. Despite the high memory usage for storing all the $NL$ embeddings, we find their strategy very useful and in practice, even for a small $r$ (e.g., 2) can lead to good convergence. 

We summarize the time and space complexity in Table~\ref{tab:complexity}. Clearly, all the SGD-based algorithms suffer from exponential complexity with respect to the number of layers, and for VR-GCN, even though $r$ can be small, they incur huge space complexity that could go beyond a GPU's memory capacity. 
In the following, we introduce our Cluster-GCN algorithm, which achieves the best of two worlds---the same time complexity per epoch with full gradient descent and the same memory complexity with vanilla SGD. 

%To formally introduce the node expansion problem, we define a new concept, ``nodes' utilization'', which counts the number of nodes searched for when computing the loss function. For vanilla mini-batch training, for compute the loss for one node, we need to search for $d^L$ nodes. Therefore the nodes' utilization is $O(Nd^L)$ per epoch if the GCN has $L$ layers and average degree for each node is $d$. We can see that node utilization for the vanilla mini-batch training is very low, and thus training time for GCN is quite heavy. 
%{\color{red}Cho: maybe modify the definition of nodes' uytilization}

\subsection{Vanilla Cluster-GCN}

Our Cluster-GCN technique is motivated by the following question: In mini-batch SGD updates, can we design a batch and the corresponding computation subgraph  to maximize the embedding utilization? 
We answer this affirmative by connecting the concept of embedding utilization to a clustering objective. 

Consider the case that in each batch we compute the embeddings for a set of nodes $\B$ from layer $1$ to $L$.  %At each layer, the embedding of node $i$ will be reused $|B_i|$ times, where $|B_i|$ is number of neighbors within set $B$. 
Since the same subgraph $A_{\B, \B}$ (links within  $\B$) is used for each layer of computation, we can then see that embedding utilization is the number of edges within this batch 
$ \|A_{\B, \B}\|_0$.  Therefore, to maximize embedding utilization, we should design a batch $\B$ to maximize the within-batch edges, by which we connect the efficiency of SGD updates with graph clustering algorithms.
 
 %To reduce node expansion and improve the utilization of each computed embedding,  one efficient way is to constraint each node to only expand inside a small area, so that when growing the layers, the number of nodes searched for will remain the same and we will be able to get rid of the exponentially growing neighborhood expansion over  layers. 

Now we formally introduce Cluster-GCN. For a graph $G$, we partition its nodes into $c$ groups: $\mathcal{V} = [\mathcal{V}_1, \cdots \mathcal{V}_c]$ where $\mathcal{V}_t$ consists of the nodes in the $t$-th partition. Thus we have $c$ subgraphs as \[\bar{G} =[G_1,\cdots, G_c]=[\{\mathcal{V}_1,\mathcal{E}_1\},\cdots, \{\mathcal{V}_c,\mathcal{E}_c\}],  \] 
where each $\mathcal{E}_t$ only consists of the links between nodes in $\mathcal{V}_t$. 
%We use $\bar{A}$ to denote the corresponding adjancency matrix of $\bar{G}$. 
%For $\bar{G}$, we will have it corresponding adjacency matrix $\bar{A}$ based on $A$.
%
%More specifically, by partitioning $N$ nodes into $c$ partitions, $[\mathcal{V}_1, \cdots \mathcal{V}_c]$ and
After reorganizing nodes, the adjacency matrix is partitioned into $c^2$ submatrices as 
\begin{equation}
\label{eq:matrix-cluster}
A =\bar{A}+\Delta = \begin{bmatrix}
A_{11} & \cdots & A_{1c} \\
\vdots & \ddots & \vdots \\
A_{c1} & \cdots & A_{cc} \end{bmatrix}
\end{equation}
and 
\begin{equation}
\bar{A} = \begin{bmatrix}
A_{11} & \cdots & 0\\
\vdots & \ddots & \vdots \\
0& \cdots & A_{cc} \end{bmatrix},
\Delta = \begin{bmatrix}
0 & \cdots &A_{1c}\\
\vdots & \ddots & \vdots \\
A_{c1}& \cdots & 0 \end{bmatrix},
\end{equation}
where each diagonal block $A_{tt}$ is a  $|\mathcal{V}_t|\times |\mathcal{V}_t|$ adjacency matrix containing the links within $G_t$. $\bar{A}$ is the adjacency matrix for graph $\bar{G}$; $A_{st}$ contains the links between two partitions $\mathcal{V}_s$ and $\mathcal{V}_t$; $\Delta$ is the matrix consisting of all off-diagonal blocks of $A$. Similarly, we can partition the feature matrix $X$ and training labels $Y$ according to the partition $[\mathcal{V}_1,\cdots, \mathcal{V}_c]$ as $[X_1,\cdots, X_c]$ and $[Y_1,\cdots, Y_c]$ where $X_t$ and $Y_t$ consist of the features and labels for the nodes in $V_t$ respectively.

The benefit of this block-diagonal approximation $\bar{G}$ is that the objective function of GCN becomes decomposible into different batches (clusters). Let $\bar{A}'$ denotes the normalized version of $\bar{A}$, the final embedding matrix becomes
\begin{align}
  Z^{(L)} &= \bar{A}' \sigma(\bar{A}'\sigma(\cdots \sigma(\bar{A}'XW^{(0)}) W^{(1)}) \cdots) W^{(L-1)}  \label{eq:decompose_embedding}\\
  &= \begin{bmatrix}
\bar{A}'_{11} \sigma(\bar{A}'_{11}\sigma(\cdots \sigma(\bar{A}'_{11}X_1W^{(0)}) W^{(1)}) \cdots) W^{(L-1)} \nonumber\\
\vdots  \nonumber\\
\bar{A}'_{cc} \sigma(\bar{A}'_{cc}\sigma(\cdots \sigma(\bar{A}'_{cc}X_cW^{(0)}) W^{(1)}) \cdots)W^{(L-1)}
%\left( \hat{A}_{cc} \ \sigma \left(\hat{A}_{cc} X_c W^{(0)}\right) \ W^{(1)} \right) 
\end{bmatrix} \nonumber
\end{align}
due to the block-diagonal form of $\bar{A}$ (note that $\bar{A}_{tt}'$ is the corresponding diagonal block of $\bar{A}'$). 
The loss function can also be decomposed into
\begin{equation}
\mathcal{L}_{\bar{A}'} = \sum_t \frac{|\mathcal{V}_t|}{N} \mathcal{L}_{\bar{A}_{tt}'} \  \ \text{ and } \ \
    \mathcal{L}_{\bar{A}_{tt}'} = \frac{1}{|\mathcal{V}_t|} \sum_{i\in \mathcal{V}_t} \text{loss}(y_i, z^{(L)}_i).
    \label{eq:decompose_obj}
\end{equation}

The Cluster-GCN is then based on the decomposition form in \eqref{eq:decompose_embedding} and \eqref{eq:decompose_obj}. At each step, we sample a cluster $\mathcal{V}_t$ and then conduct SGD to update based on the gradient of $\mathcal{L}_{\bar{A}_{tt}'}$, 
and this only requires the sub-graph $A_{tt}$, the  $X_t$, $Y_t$ on the current batch and the models $\{W^{(l)}\}_{l=1}^L$. The implementation only requires forward and backward propagation of matrix products (one block of \eqref{eq:decompose_embedding}) that is much easier to implement than the neighborhood search procedure used in previous SGD-based training methods.

%In general, we can define the loss function for each subgraph as 
%\begin{equation*}
%    \mathcal{L}_{A_{tt}} = \frac{1}{|\mathcal{V}_t|} \sum_{i\in \mathcal{V}_t} \text{loss}(y_i, Z^{(L)}_i)
%\end{equation*}
%and since $Z^{(L)}$ only 

%To train GCN using mini-batch on $\bar{G}$, we construct each batch $\bar{B}$ consists of nodes in $\mathcal{V}_i$. When evaluating 
%\begin{align}
%   f(X, \bar{A}) 
%   &= \text{softmax} \left( \hat{A} \ \text{ReLU} \left(\hat{A} X W^{(0)}\right) \ W^{(1)} \right) \\ \nonumber
%   &=\text{softmax} \begin{bmatrix}
%\left( \hat{A}_{11} \ \text{ReLU} \left(\hat{A}_{11} X_1 W^{(0)}\right) \ W^{(1)} \right)  \\
%\vdots  \\
%\left( \hat{A}_{cc} \ \text{ReLU} \left(\hat{A}_{cc} X_c W^{(0)}\right) \ W^{(1)} \right) 
%\end{bmatrix}
%\end{align}
%we only need to load $A_{ii}, X_i,Y_i$, and the models $\{W_i\}_{i=1}^L$. So the node partition will be memory efficient.

We use graph clustering algorithms to partition the graph. 
Graph clustering methods such as Metis~\cite{Metis} and Graclus~\cite{Graclus} aim to construct the partitions over the vertices in the graph such that within-clusters links are much more than between-cluster links to better capture the clustering and community structure of the graph. These are exactly what we need because: 
1) As mentioned before, the embedding utilization is equivalent to the within-cluster links for each batch. Intuitively, each node and its neighbors are usually located in the same cluster, therefore after a few hops, neighborhood nodes with a high chance are still in the same cluster. 
2) Since we replace $A$ by its block diagonal approximation $\bar{A}$ and the error is proportional to between-cluster links $\Delta$, we need to find a partition to minimize number of  between-cluster links. 
%How to partition the nodes will be challenging as if the partitions are constructed in a way that too many edges are cut in the graph $\bar{G}$, it will break too many connections among partitions and hurt the final accuracy. 
%On the other hand, graph represents the relationship between two nodes, and usually exhibits clustering structure. 
 
 In Figure~\ref{fig:cluster}, we illustrate the neighborhood expansion with full graph $G$ and the graph with clustering partition $\bar{G}$. 
 We can see that cluster-GCN can avoid heavy neighborhood search and focus on the neighbors within each cluster. 
 %By using clustering batch, we can improve the node utilization and thus speed up the training. 
 In Table \ref{tab:random}, we show two different node partition strategies:  random partition versus clustering partition.
 We partition the graph into 10 parts by using random partition and METIS.
 Then use one partition as a batch to perform a SGD update. 
 %We then use mini-batch based the new graph, where we construct each batch using one cluster's node and their links inside each cluster for the same number for epochs for two partitions. 
 We can see that with the same number of epochs, using clustering partition can achieve higher accuracy. This shows using graph clustering is important and partitions should not be formed randomly. 
 
\paragraph{\bf Time and space complexity. }
Since each node in $\mathcal{V}_t$ only links to nodes inside $\mathcal{V}_t$, each node does not need to perform neighborhoods searching outside $A_{tt}$. %$A_{tt}$ will include all the hop-$l$ ($l=1,\cdots, L$) neighbors for each node in $\mathcal{V}_t$ and it is sufficient for convolution over layers.
The computation for each batch will purely be matrix products $\bar{A}_{tt}' X^{(l)}_t W^{(l)}$ and some element-wise operations, so the overall time complexity per batch is $O(\|{A}_{tt}\|_0 F + bF^2)$.
Thus the overall time complexity per epoch becomes $O(\|A\|_0 F + NF^2)$. 
In average, each batch only requires computing $O(bL)$ embeddings, which is linear instead of exponential to $L$. 
%The node utilization is thus $O(NdL)$. Compared with the vanilla mini-batch training for GCN,  which has node utilization $O(Nd^L)$, this block approximation can significantly improve the training time. 
In terms of space complexity, in each batch, we only need to load $b$ samples and store their embeddings on each layer, resulting in $O(bLF)$ memory for storing embeddings.
Therefore our algorithm is also more memory efficient than all the previous algorithms. 
Moreover, our algorithm only requires loading a subgraph into GPU memory instead of the full graph (though graph is usually not the memory bottleneck). 
%\footnote{Note that in practice GCN implementations usually load full graph into GPU memory since it is relatively small compared with the embedding and feature parts. {\color{red}Cho: Wei-lin is this correct?}}. 
The detailed time and memory complexity are summarized in Table~\ref{tab:complexity}. 
%need to load much larger graph in $G$ which contains all the neighbors for nodes $\mathcal{V}_i$ and their hop-$l$ neighbors outside $A_ii$, sometime can even grow to the whole graph $G$.

\begin{table}
\caption{Random partition versus clustering partition of the graph (trained on mini-batch SGD). Clustering partition leads to better performance (in terms of test F1 score) since it removes less between-partition links. These three datasetes are all public GCN datasets. We will explain PPI data in the experiment part. Cora has 2,708 nodes and 13,264 edges, and Pubmed has 19,717 nodes and 108,365 edges. }
  \centering
  \renewcommand\bfdefault{b}
  \vspace{-10pt}\begin{tabular}{|c|c|c|}
  \hline
   Dataset &  random partition  & clustering partition \\
  \hline
  Cora &  78.4 & 82.5 \\
  \hline
  Pubmed &  78.9  & 79.9 \\
  \hline
  PPI &   68.1 & 92.9\\
\hline
\end{tabular}
\label{tab:random}
\vspace{-10pt}
\end{table}

\subsection{Stochastic Multiple Partitions}
\label{sec:multiple}
\begin{figure}
 \centering 
    \includegraphics[width=.7\linewidth]{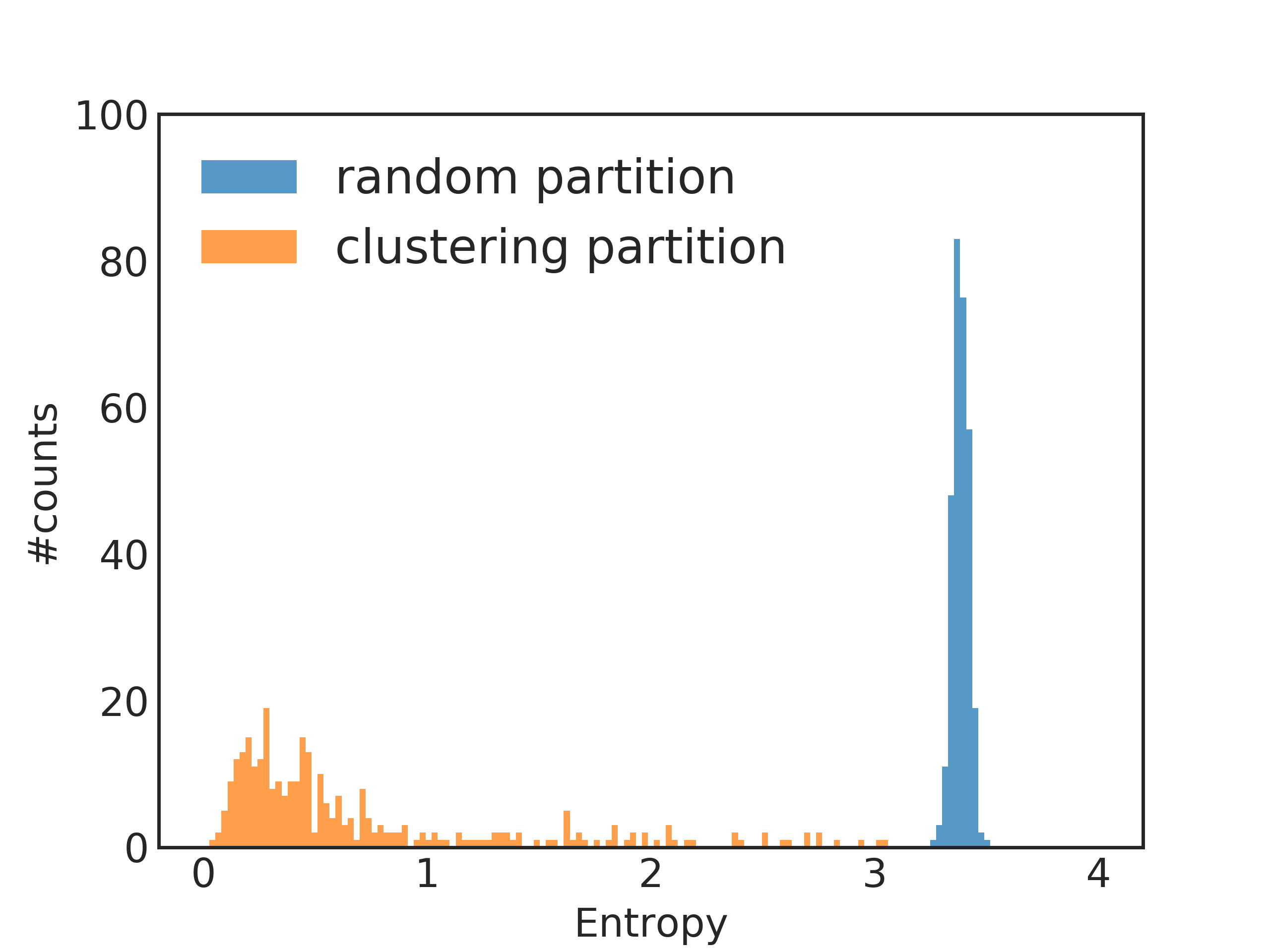}
    \caption{Histograms of entropy values based on the label distribution. Here we present  
    within each batch using random partition versus clustering partition. Most clustering partitioned batches have low label entropy, indicating skewed label distribution within each batch. In comparison, random partition will lead to larger label entropy within a batch although it is less efficient as discussed earlier. We partition the Reddit dataset with 300 clusters in this example.  \label{fig:histogram}}
    \vspace{-10pt}
\end{figure}

Although vanilla Cluster-GCN achieves good computational and memory complexity, there are still two potential issues: 
%However as shown in Table \ref{tab:random}, there is still some accuracy difference between the clustering partitions of graph with mini-batch and using full-batch over original graph, although each epoch the former one is much faster and memory efficient than the later one. There are two reasons for that:
\begin{itemize}
    \item After the graph is partitioned, some links (the $\Delta$ part in Eq.~\eqref{eq:matrix-cluster}) are removed. Thus the performance could be affected.
    \item Graph clustering algorithms tend to bring similar nodes together. Hence the distribution of a cluster could be different from the original data set, leading to a biased estimation of the full gradient while performing SGD updates.
\end{itemize}
In Figure~\ref{fig:histogram}, we demonstrate an example of unbalanced label distribution by using the Reddit data with clusters formed by Metis.
We calculate the entropy value of each cluster based on its label distribution.
Comparing with random partitioning, we clearly see that entropy of most clusters are smaller, indicating that the label distributions of clusters are biased towards some specific labels.
This increases the variance across different batches and may affect the convergence of SGD. 
%This in other hand will increase the variance of mini-batch SGD. 
%bring the variance issue for clustering partition.  

% To conquer the above issues, one solution is to construct multiple graph partitions, and at each epoch follow  one partition to choose batches. In this way, each link in the original graph have some chance to be considered in different epochs. However,  multiple graph partition is less practical due to the memory overhead and the hardness of obtaining many different clustering partitions. 

\begin{figure}
 \centering 
    \includegraphics[width=1.0\linewidth]{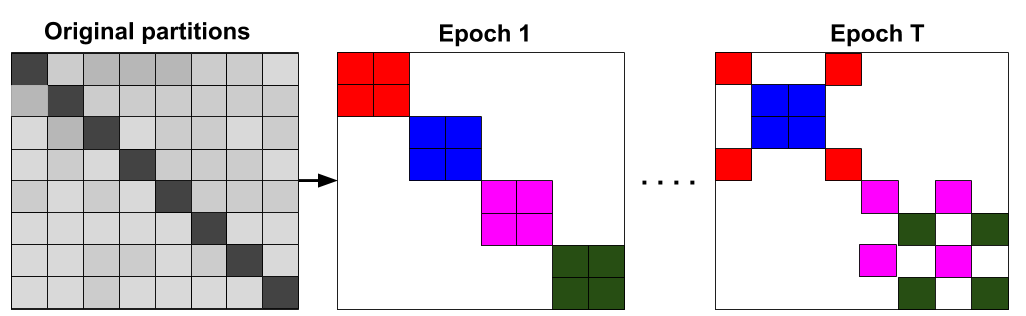}
    \caption{The proposed stochastic multiple partitions scheme. In each epoch, we randomly sample $q$ clusters ($q=2$ is used in this example) and their between-cluster links to form a new batch. Same color blocks are in the same batch. }
    \label{fig:cluster2}
    \vspace{-10pt}
\end{figure}

\begin{figure}
 \centering 
    \includegraphics[width=0.7\linewidth]{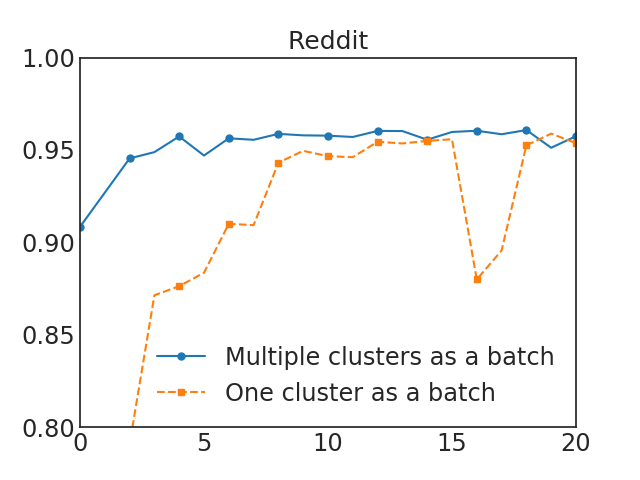}
    \caption{Comparisons of choosing one cluster versus multiple clusters. The former uses 300 partitions. The latter uses 1500 and randomly select 5 to form one batch. We present epoch (x-axis) versus F1 score (y-axis).}
    \label{fig:multiple}
    \vspace{-10pt}
\end{figure}

To address the above issues, we propose a stochastic multiple clustering approach to incorporate between-cluster links and reduce variance across batches.
We first partition the graph into $p$ clusters $\mathcal{V}_1,\cdots, \mathcal{V}_p$ with a relatively large $p$.
When constructing a batch $B$ for an SGD update, instead of considering only one cluster, we randomly choose $q$ clusters, denoted as $t_1, \dots, t_q$ and include their nodes $\{\mathcal{V}_{t_1} \cup \dots \cup \mathcal{V}_{t_q}\}$ into the batch.
Furthermore, the links between the chosen clusters, 
\begin{equation*}
    \{A_{ij} \mid i,j \in t_1, \dots, t_q\},
\end{equation*}
are added back.
% Furthermore, in addition to the diagonal blocks of $A$, we include all the between-cluster links within these $q$ clusters: $\{A_{pq} \mid p,q = t_1, \dots, t_q \}$ into the subgraph of the batch. 
In this way, those between-cluster links are re-incorporated and the combinations of clusters make the variance across batches smaller.
% and we will include different sets of links and nodes that are partitioned into different clusters in each epoch, which addresses the issue of missing links and increasing the variance within each batch---that helps the convergence. 
Figure~\ref{fig:cluster2} illustrates our algorithm---for each epochs, different combinations of clusters are chosen as a batch. 
We conduct an experiment on Reddit to demonstrate the effectiveness of the proposed approach.
In Figure~\ref{fig:multiple}, we can observe that using multiple clusters as one batch could improve the convergence.
Our final Cluster-GCN algorithm is presented in Algorithm~\ref{alg:main1}.
% By alternating the partitions over different epochs, we will have different graph partitions. Since all the different partitions are generated from the same clustering, we can store them as blocks of subgraphs and in different epoch, we can combine different blocks to generate new partitions. As a consequence, we do not need additional memory to store these graphs even though different graph structures are used in different epochs. Figure\ref{fig:cluster2} illustrates our algorithm---for each epochs, different clusters are grouped into one bigger partition as a batch. 

%{\color{red}(Cho: Add a comparison between vanilla cluster-GCN and Stochastic Multiple clusters here. )} 
 
\begin{algorithm}[t]
\label{alg:main1}
\caption{Cluster GCN}
 \label{alg:main1}
 \KwIn{Graph $A$, feature $X$, label $Y$; }
 \KwOut{Node representation $\bar{X}$}
Partition graph nodes into $c$ clusters $\mathcal{V}_1,\mathcal{V}_2,\cdots,\mathcal{V}_c$ by METIS\;
\For{$\text{iter}=1,\cdots, \text{max\_iter}$}{
Randomly choose $q$ clusters, $t_1, \cdots, t_q$ from $\mathcal{V}$ without replacement\;
Form the subgraph $\bar{G}$ with nodes $\bar{\mathcal{V}} = [\mathcal{V}_{t_1},\mathcal{V}_{t_2},\cdots,\mathcal{V}_{t_q}]$ and links $A_{\bar{\mathcal{V}}, \bar{\mathcal{V}}}$ \;
%within $[\mathcal{V}_{t_1},\mathcal{V}_{t_2},\cdots,\mathcal{V}_{t_q}]$\;
Compute $g \leftarrow \nabla \mathcal{L}_{A_{\bar{\mathcal{V}}, \bar{\mathcal{V}}}}$ (loss on the subgraph $A_{\bar{\mathcal{V}}, \bar{\mathcal{V}}}$) \;
Conduct Adam update using gradient estimator $g$
} 

Output: $\{W_l\}_{l=1}^L$
\end{algorithm} 

\subsection{Issues of training deeper GCNs}

Previous attempts of training deeper GCNs \cite{kipf2017semi} seem to suggest that adding more layers is not helpful. However, the datasets used in the experiments may be too small to make a proper justification.
For example, \cite{kipf2017semi} considered a graph with only a few hundreds of training nodes for which overfitting can be an issue.
%When considering large datasets, scalability is another challenging one.
Moreover, we observe that the optimization of deep GCN models becomes difficult as it may impede the information from the first few layers being passed through.
In \cite{kipf2017semi}, they adopt a technique similar to residual connections \cite{he2016deep} to enable the model to carry the information from a previous layer to a next layer.
Specifically, they modify \eqref{eq:gcn0} to add the hidden representations of layer $l$ into the next layer.
\begin{equation}
\label{eq:residual}
    X^{(l+1)} = \sigma(A' X^{(l)} W^{(l)}) + X^{ (l)}
\end{equation}
Here we propose another simple technique to improve the training of deep GCNs.
In the original GCN settings, each node aggregates the representation of its neighbors from the previous layer.
However, under the setting of deep GCNs, the strategy may not be suitable as it does not take the number of layers into account.
Intuitively, neighbors nearby should contribute more than distant nodes.
We thus propose a technique to better address this issue.
The idea is to amplify the diagonal parts of the adjacency matrix $A$ used in each GCN layer.
In this way, we are putting more weights on the representation from the previous layer in the aggregation of each GCN layer.
An example is to add an identity to $\bar{A}$ as follows.
\begin{equation}
\label{eq:diag-enhancement}
    X^{(l+1)} = \sigma((A' + I) X^{(l)} W^{(l)})
\end{equation}
While \eqref{eq:diag-enhancement} seems to be reasonable, using the same weight for all the nodes regardless of their numbers of neighbors may not be suitable.
Moreover, it may suffer from numerical instability as values can grow exponentially when more layers are used.
Hence we propose a modified version of  \eqref{eq:diag-enhancement} to better maintain the neighborhoods information and numerical ranges.
We first add an identity to the original $A$ and perform the normalization
\begin{equation}
\label{eq:normalization-id}
    \tilde{A} = (D+I)^{-1} (A+I),
\end{equation} and then consider
\begin{equation}
\label{eq:diag-enhancement-modified}
    X^{(l+1)} = \sigma((\tilde{A} + \lambda\text{diag}(\tilde{A})) X^{(l)} W^{(l)}).
\end{equation}
Experimental results of adopting the ``diagonal enhancement'' techniques are presented in Section~\ref{sec:deepgcn} where we show that this new normalization strategy can help to build deep GCN and achieve SOTA performance.

% In the earlier experiments, the following normalized adjacency matrix is used
% \begin{equation}
% \label{eq:normalization}
%     A' = D^{-1} A,
% \end{equation}
% where $D$ is a diagonal matrix with $D_{ii} = \sum_{j} A_{ij}$.
% Another normalization technique is to add an identity to the original $A$ and we have

%As to enhance the diagonal elements of $A$, we also include another normalization on $A$ with an identity added.

% \subsection{Implementation details}

% Implementing a graph convolution network that works on large graphs involves several details.
% Firstly, previous works \citep{chen2018fastgcn, chen2018stochastic} propose to pre-compute the multiplication of $AX$ in the first GCN layer.
% %This strategy indeed gives lots of benefits in terms of accuracy and time.
% By precomputing $AX$, we are essentially using the exact 1-hop neighborhood for each node and the expensive neighbors searching in the first layer can be saved.
% In Table ? we present the running time and accuracy of methods with and without the preprocess strategy.

% Renormalization?

\section{Experiments}
\label{sec:exp}

We evaluate our proposed method for training GCN on two tasks: multi-label and multi-class classification on four public datasets. 
The statistic of the data sets are shown in Table \ref{tab:data}.
Note that the Reddit dataset is the largest public dataset we have seen so far for GCN, and the Amazon2M dataset is collected by ourselves and is much larger than Reddit (see more details in Section~\ref{sec:amazon}). 

%For multi-class classification, we evaluate on Reddit data set (one largest public data set we have seen so far for GCN) and Amazon2M. For multi-label classification task, we evaluate on PPI and Amazon. 
     \begin{table}
\centering
      \caption{Data statistics\label{tab:data}}
    \vspace{-10pt}
    \resizebox{0.48 \textwidth}{!}{
    \begin{tabular}{|l|c|r|r|r|r|}
    \hline
    Datasets & Task & \#Nodes & \#Edges & \#Labels & \#Features \\
%     \hline
     \hline
     PPI & multi-label & 56,944 & 818,716 & 121 & 50  \\
     \hline
     Reddit & multi-class & 232,965 & 11,606,919 & 41 & 602 \\
     \hline
     Amazon & multi-label & 334,863 & 925,872 & 58 & N/A  \\
     \hline %91973/28285/214605
   Amazon2M & multi-class & 2,449,029 & 61,859,140 & 47 & 100  \\
   \hline %1709997/739032
    \end{tabular}   }
  \end{table}

\begin{table}
\centering
      \caption{The parameters used in the experiments.}
      \label{tab:parameter}
      \vspace{-10pt}
      \resizebox{0.4 \textwidth}{!}{
    \begin{tabular}{|l|r|r|r|}
    \hline
    Datasets & \#hidden units & \# partitions & \#clusters per batch \\
     \hline
     PPI  & 512 & 50 & 1\\
     \hline
     Reddit  & 128 & 1500 & 20 \\
     \hline
     Amazon  & 128 & 200 & 1\\
     \hline %91973/28285/214605
  Amazon2M  & 400 & 15000 & 10\\
  \hline %1709997/739032
    \end{tabular} }
    \vspace{-10pt}
  \end{table}

We include the following state-of-the-art GCN training algorithms in our comparisons: 
\begin{itemize}
\item Cluster-GCN (Our proposed algorithm): the proposed fast GCN training method. 
%(Maybe mention how we select parameters to form the curve?)}
\item VRGCN\footnote{GitHub link: \url{https://github.com/thu-ml/stochastic_gcn}}~\cite{chen2018stochastic}: It maintains the historical embedding of all the nodes in the graph and expands to only a few neighbors to speedup training. The number of sampled neighbors is set to be 2 as suggested in \cite{chen2018stochastic}\footnote{Note that we also tried the default sample size 20 in VRGCN package but it performs much worse than sample size$=2$. }.
\item GraphSAGE\footnote{GitHub link: \url{https://github.com/williamleif/GraphSAGE}}~\citep{hamilton2017inductive}: It samples a fixed number of neighbors per node. We use the default settings of sampled sizes for each layer ($S_1=25, S_2=10$) in GraphSAGE.
% \item FastGCN\footnote{GitHub link: \url{https://github.com/matenure/FastGCN}}~\cite{chen2018fastgcn}: It considers importance sampling and chooses a fixed number of neighbors in each layer.
\end{itemize}
We implement our method in PyTorch \cite{paszke2017automatic}. For the other methods, we use all the original papers' code from their github pages. Since \cite{kipf2017semi} has difficulty to scale to large graphs, we do not compare with it here.
Also as shown in \cite{chen2018stochastic} that VRGCN is faster than FastGCN, so we do not compare with FastGCN here.
For all the methods we use the Adam optimizer with learning rate as 0.01, dropout rate as 20\%, weight decay as zero.
The mean aggregator proposed by \cite{hamilton2017inductive} is adopted and the number of hidden units is the same for all methods.
Note that techniques such as \eqref{eq:diag-enhancement-modified} is not considered here.
In each experiment, we consider the same GCN architecture for all methods.
For VRGCN and GraphSAGE, we follow the settings provided by the original papers and set the batch sizes as 512.
For Cluster-GCN, the number of partitions and clusters per batch for each dataset are listed in Table~\ref{tab:parameter}.
Note that clustering is seen as a preprocessing step and its running time is not taken into account in training.
In Section~\ref{sec:more-details}, we show that graph clustering only takes a small portion of preprocessing time.
All the experiments are conducted on a machine with a NVIDIA Tesla V100 GPU (16 GB memory), 20-core Intel Xeon CPU (2.20 GHz), and 192 GB of RAM.

%\textcolor{red}{normalization, batch size, partition size, dropout?, machine configuration to be added}

\subsection{Training Performance for median size datasets}
\label{sec:median}

\begin{table*}
\caption{Comparisons of memory usages on different datasets. Numbers in the brackets indicate the size of hidden units used in the model.}
\vspace{-10pt}\begin{tabular}{|l|r|r|r|r|r|r|r|r|r|}
\hline
 & \multicolumn{3}{c|}{2-layer} & \multicolumn{3}{c|}{3-layer} & \multicolumn{3}{c|}{4-layer} \\ \cline{2-10} 
& VRGCN & Cluster-GCN & GraphSAGE & VRGCN & Cluster-GCN & GraphSAGE & VRGCN & Cluster-GCN & GraphSAGE\\ \hline
PPI (512)    & 258 MB  & 39 MB  & 51 MB   & 373 MB & 46 MB  & 71 MB   & 522 MB & 55 MB  & 85 MB\\ \hline
Reddit (128) & 259 MB  & 284 MB & 1074 MB & 372 MB & 285 MB & 1075 MB & 515 MB & 285 MB & 1076 MB\\ \hline
Reddit (512) & 1031 MB & 292 MB & 1099 MB & 1491 MB & 300 MB & 1115 MB & 2064 MB & 308 MB & 1131 MB\\ \hline
Amazon (128) & 1188 MB & 703 MB & N/A & 1351 MB & 704 MB & N/A & 1515 MB & 705 MB & N/A \\ \hline
\end{tabular}
\label{tab:memory}
\end{table*}

{\bf Training Time vs Accuracy:} First we compare our proposed method with other methods in terms of training speed. In Figure \ref{fig:time_acc_all}, the $x$-axis shows the training time in seconds, and $y$-axis shows the accuracy (F1 score) on the validation sets. We plot the training time versus accuracy for three datasets with 2,3,4 layers of GCN. 
% Since FastGCN is much slower than GraphSGAE, VRGCN and our method, its results cannot be seen from the figures. And 
Since GraphSAGE is slower than VRGCN and our method, the curves for GraphSAGE only appear for PPI and Reddit datasets. We can see that our method is the fastest for both PPI and Reddit datasets for GCNs with different numbers of layers.

\begin{table}
\centering
      \caption{Benchmarking on the Sparse Tensor operations in PyTorch and TensorFlow. A network with two linear layers is used and the timing includes forward and backward operations. Numbers in the brackets indicate the size of hidden units in the first layer. Amazon data is used.}
      \label{tab:benchmark}
    \vspace{-10pt}\begin{tabular}{|l|r|r|}
    \hline
        & PyTorch & TensorFlow \\
     \hline
     Avg. time per epoch (128)  & 8.81s & 2.53s\\
     \hline
     Avg. time per epoch (512)  & 45.08s & 7.13s\\
    \hline
\end{tabular} 
\vspace{-10pt}
\end{table}

For Amazon data, since nodes' features are not available, an identity matrix is used as the feature matrix $X$.
Under this setting, the shape of parameter matrix $W^{(0)}$ becomes 334863x128.
Therefore, the computation is dominated by sparse matrix operations such as $AW^{(0)}$. Our method is still faster than VRGCN for 3-layer case, but slower for 2-layer and 4-layer ones. The reason may come from the speed of sparse matrix operations from different frameworks.
VRGCN is implemented in TensorFlow, while Cluster-GCN is implemented in PyTorch whose sparse tensor support are still in its very early stage.
In Table~\ref{tab:benchmark}, we show the time for TensorFlow and PyTorch to do forward/backward operations on Amazon data, and a simple two-layer network are used for benchmarking both frameworks.
We can clearly see that TensorFlow is faster than PyTorch. The difference is more significant when the number of hidden units increases.
This may explain why Cluster-GCN has longer training time in Amazon dataset.
%Therefore, Cluster-GCN is slightly worse than VRGCN in terms of overall training time due to slower matrix multiplication in PyTorch. 

{\bf Memory usage comparison:} For training large-scale GCNs, besides training time, memory usage needed for training is often more important and will directly restrict the scalability. The memory usage includes the memory needed for training the GCN for many epochs. 
%Since VRGCN is much faster than GraphSAGE, we only compare our memory usage with VRGCN. 
As discussed in Section~\ref{sec:algorithm}, to speedup training, VRGCN needs to save historical embeddings during training, so it needs much more memory for training than Cluster-GCN. GraphSAGE also has higher memory requirement than Cluster-GCN due to the exponential neighborhood growing problem. In Table \ref{tab:memory}, we compare our memory usage with VRGCN's memory usage for GCN with different layers. When increasing the number of layers, Cluster-GCN's memory usage does not increase a lot. The reason is that when increasing one layer, the extra variable introduced is the weight matrix $W^{(L)}$, which is relatively small comparing to the sub-graph and node features. While VRGCN needs to save each layer's history embeddings, and the embeddings are usually dense and will soon dominate the memory usage. We can see from Table \ref{tab:memory} that Cluster-GCN is much more memory efficient than VRGCN. For instance, on Reddit data to train a 4-layer GCN with hidden dimension to be 512, VRGCN needs 2064MB memory, while Cluster-GCN only uses 308MB memory.

\begin{table}
\caption{The most common categories in Amazon2M.}
  \centering
  \renewcommand\bfdefault{b}
  \vspace{-10pt}\begin{tabular}{|c|c|}
  \hline
  Categories & number of products\\
  \hline
  Books & 668,950\\
  \hline
  CDs \& Vinyl & 172,199\\
  \hline
  Toys \& Games & 158,771\\
\hline
%\news & 19,996 & 1,355,191 & 0.03\% & ? \\
\end{tabular}
\label{tab:category}
%\vskip -0.1in
\vspace{-10pt}
\end{table}

\subsection{Experimental results on Amazon2M}
\label{sec:amazon}
\begin{figure*}[tb]
    \begin{tabular}{@{}l@{}l@{}l@{}}
    %vspace{-10px}
    \subfigure[PPI (2 layers)]{\includegraphics[width=.33\linewidth,trim=0mm 0mm 0mm 0mm,clip=true]{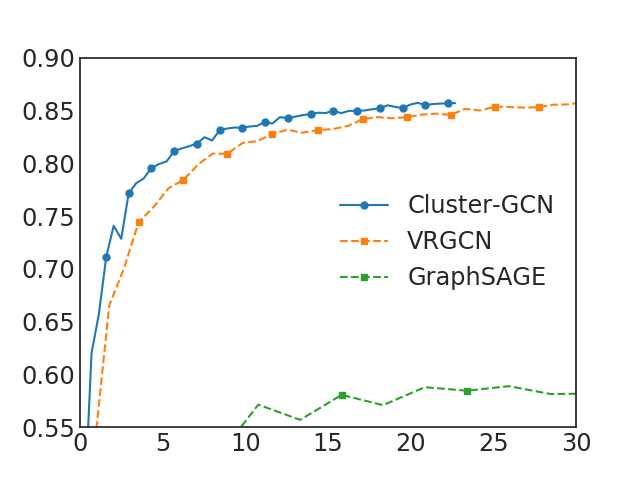}} &
    \subfigure[PPI (3 layers)]{\includegraphics[width=.33\linewidth,trim=0mm 0mm 0mm 0mm,clip=true]{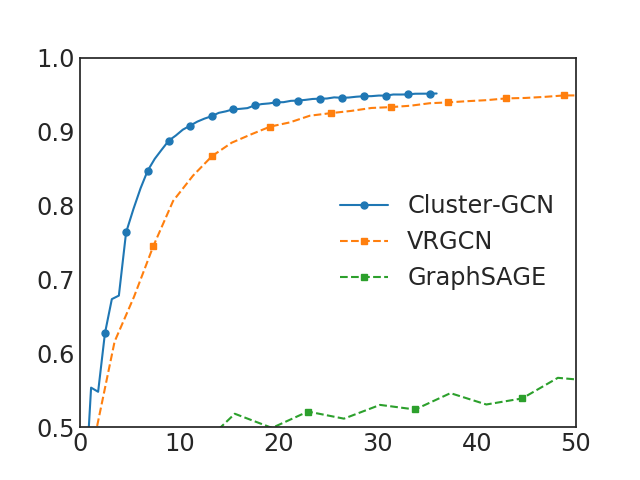}} &
    \subfigure[PPI (4 layers)]{\includegraphics[width=.33\linewidth,trim=0mm 0mm 0mm 0mm,clip=true]{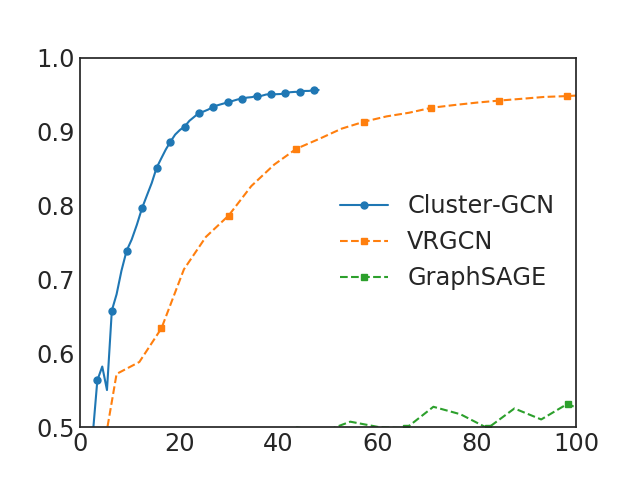}}  \\
    \subfigure[Reddit (2 layers)]{\includegraphics[width=.33\linewidth,trim=0mm 0mm 0mm 0mm,clip=true]{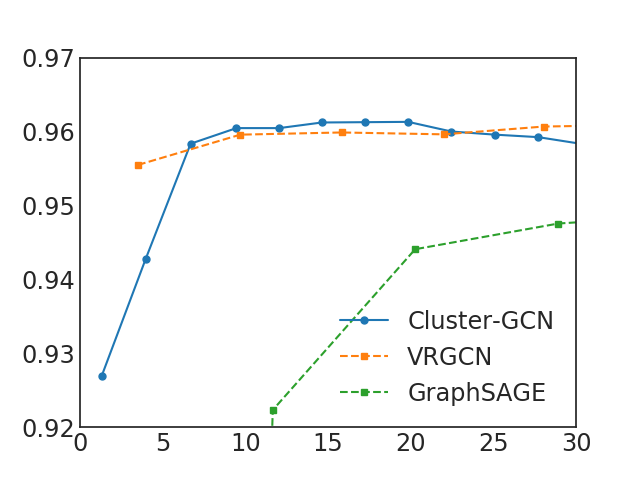}} &
    \subfigure[Reddit (3 layers)]{\includegraphics[width=.33\linewidth,trim=0mm 0mm 0mm 0mm,clip=true]{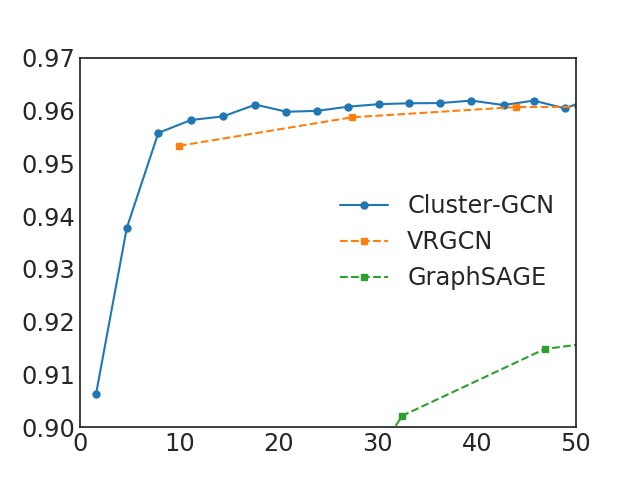}} &
    \subfigure[Reddit (4 layers)]{\includegraphics[width=.33\linewidth,trim=0mm 0mm 0mm 0mm,clip=true]{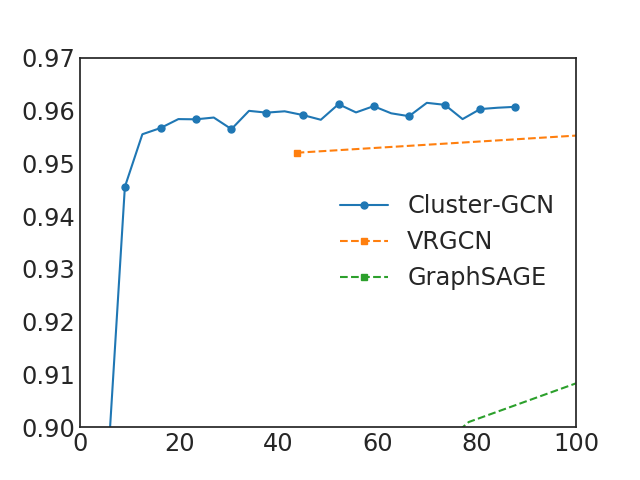}}  \\
    \subfigure[Amazon (2 layers)]{\includegraphics[width=.33\linewidth,trim=0mm 0mm 0mm 0mm,clip=true]{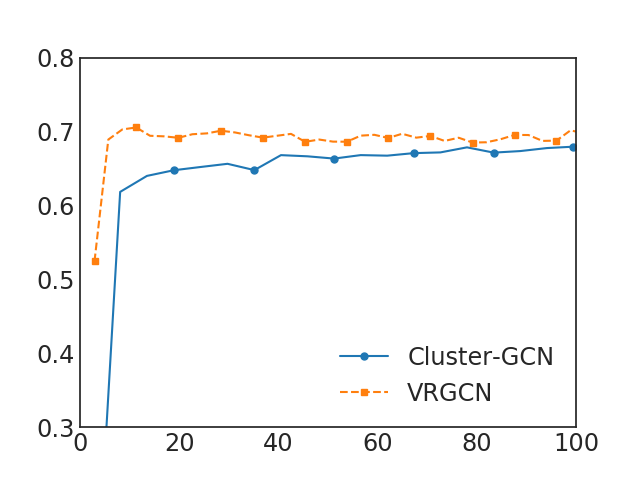}} &
    \subfigure[Amazon (3 layers)]{\includegraphics[width=.33\linewidth,trim=0mm 0mm 0mm 0mm,clip=true]{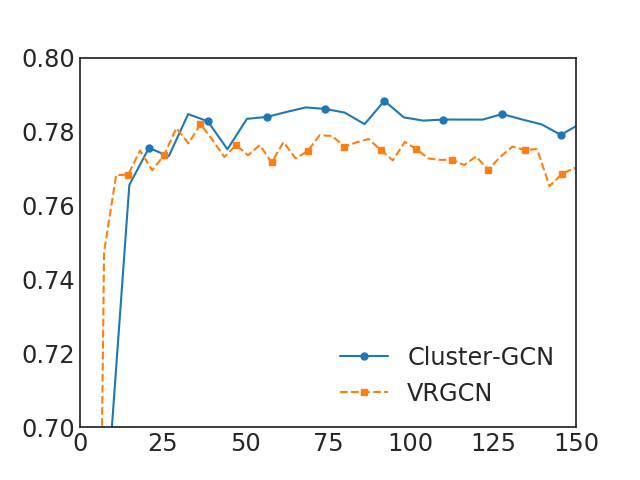}} &
    \subfigure[Amazon (4 layers)]{\includegraphics[width=.33\linewidth,trim=0mm 0mm 0mm 0mm,clip=true]{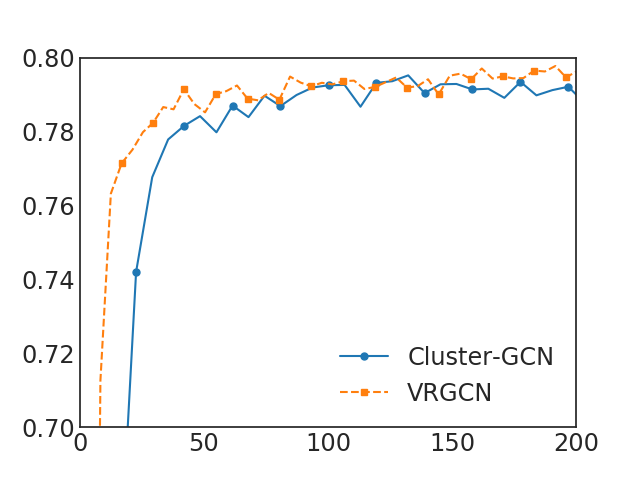}}  \\
    %\vspace{-10px}
    \end{tabular}
    \caption{Comparisons of different GCN training methods. We present the relation between training time in seconds (x-axis) and the validation F1 score (y-axis).}
    \label{fig:time_acc_all}
    %\vspace{30pt}
\end{figure*}

{\bf A new GCN dataset: Amazon2M.}
By far the largest public data for testing GCN is Reddit dataset with the statistics shown in Table \ref{tab:data}, which contains about 200K nodes. As shown in Figure \ref{fig:time_acc_all} GCN training on this data can be finished within a few hundreds seconds.
To test the scalability of GCN training algorithms, we constructed a much larger graph with over 2 millions of nodes and 61 million edges based on Amazon co-purchasing networks \citep{mcauley2015image,mcauley2015inferring}.  The raw co-purchase data is from \textsc{Amazon-3M}\footnote{\url{http://manikvarma.org/downloads/XC/XMLRepository.html}}. In the graph, each node is a product, and the graph link represents whether two products are purchased together. Each node feature is generated by extracting bag-of-word features from the product descriptions followed by Principal Component Analysis~\cite{PCA} to reduce the dimension to be 100. In addition, we use the top-level categories as the labels for that product/node (see Table \ref{tab:category} for the most common categories). The detailed statistics of the data set are listed in Table \ref{tab:data}.

In Table \ref{tab:amazon2m}, we compare with VRGCN for GCNs with a different number of layers in terms of training time, memory usage, and test accuracy (F1 score). As can be seen from the table that 1) VRGCN is faster than Cluster-GCN with 2-layer GCN but slower than Cluster-GCN when increasing one layer while achieving similar accuracy. 2) In terms of memory usage, VRGCN is using much more memory than Cluster-GCN (5 times more for 3-layer case), and it is running out of memory when training 4-layer GCN, while Cluster-GCN does not need much additional memory when increasing the number of layers, and achieves the best accuracy for this data when training a 4-layer GCN. 

\begin{table*}
\caption{Comparisons of running time, memory and testing accuracy (F1 score) for Amazon2M.}
\vspace{-10pt}\begin{tabular}{|l|r|r|r|r|r|r|}
\hline
 & \multicolumn{2}{c|}{Time} & \multicolumn{2}{c|}{Memory} & \multicolumn{2}{c|}{Test F1 score} \\ \cline{2-7}
& VRGCN & Cluster-GCN & VRGCN & Cluster-GCN & VRGCN & Cluster-GCN \\
\hline
Amazon2M (2-layer) &  337s   &  1223s        &           7476 MB               &  2228 MB     & 89.03 &  89.00\\
\hline
Amazon2M (3-layer) & 1961s    & 1523s          &                      11218 MB    & 2235 MB      & 90.21         & 90.21\\
\hline
Amazon2M (4-layer) & N/A      & 2289s          & OOM & 2241 MB      & N/A           & 90.41\\
\hline
\end{tabular}
\label{tab:amazon2m}
%\vspace{pt}
\end{table*}

\subsection{Training Deeper GCN}
\label{sec:deepgcn}
%Since we have solved the accuracy issue for training deep GCNs in subsection 4.3, now we will show some comparisons for training deep GCNs in terms of training time. In Table \ref{tab:deep}, we again compare with VRGCN in terms of speed. 
%{\color{red}why we do not have accuracy here?}
In this section we consider GCNs with more layers.
We first show the timing comparisons of Cluster-GCN and VRGCN in Table \ref{tab:deep}.
PPI is used for benchmarking and we run 200 epochs for both methods.
We observe that the running time of VRGCN grows exponentially because of its expensive neighborhood finding, while the running time of Cluster-GCN only grows linearly.

\begin{figure}
 \centering 
    \includegraphics[width=0.7\linewidth]{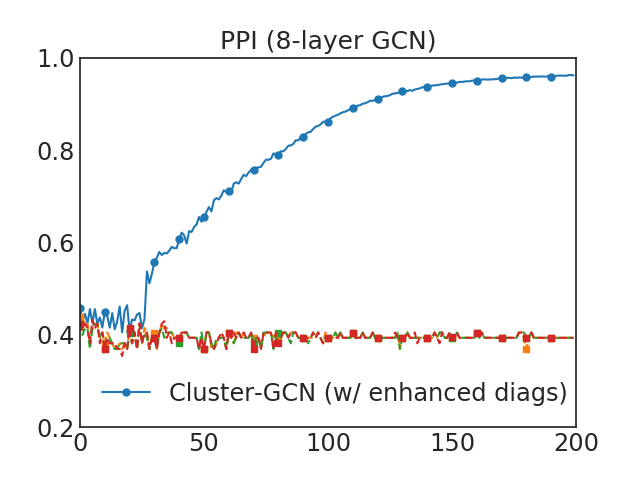}
    \caption{Convergence figure on a 8-layer GCN. We present numbers of epochs (x-axis) versus validation accuracy (y-axis). All methods except for the one using \eqref{eq:diag-enhancement-modified} fail to converge.}
    \label{fig:8layer}
    \vspace{-10pt}
\end{figure}

Next we investigate whether using deeper GCNs obtains better accuracy.
In Section~\ref{sec:deepgcn}, we discuss different strategies of modifying the adjacency matrix $A$ to facilitate the training of deep GCNs.
% In the earlier experiments, the following normalized adjacency matrix is used
% \begin{equation}
% \label{eq:normalization}
%     A' = D^{-1} A,
% \end{equation}
% where $D$ is a diagonal matrix with $D_{ii} = \sum_{j} A_{ij}$.
% Another normalization technique is to add an identity to the original $A$ and we have
% \begin{equation}
% \label{eq:normalization-id}
%     \hat{A} = (D+I)^{-1} (A+I).
% \end{equation}
% The following strategies are considered.
% \begin{itemize}
%     \item $A'$
%     \item $\hat{A}$
%     \item $\hat{A}$ + \eqref{eq:diag-enhancement}
%     \item $\hat{A}$ + \eqref{eq:diag-enhancement-modified} with $\lambda=1$
% \end{itemize}
We apply the diagonal enhancement techniques to deep GCNs and run experiments on PPI.
Results are shown in Table~\ref{tab:deep-accuracy}.
For the case of 2 to 5 layers, the accuracy of all methods increases with more layers added, suggesting that deeper GCNs may be useful.
However, when 7 or 8 GCN layers are used, the first three methods fail to converge within 200 epochs and get a dramatic loss of accuracy.
A possible reason is that the optimization for deeper GCNs becomes more difficult.
We show a detailed convergence of a 8-layer GCN in Figure~\ref{fig:8layer}.
With the proposed diagonal enhancement technique \eqref{eq:diag-enhancement-modified}, the convergence can be improved significantly and similar accuracy can be achieved. 

{\bf State-of-the-art results by training deeper GCNs. }
With the design of Cluster-GCN and the proposed normalization approach, we now have the ability for training much deeper GCNs to achieve better accuracy (F1 score). 
%With the ability of training much deeper GCNs enabled by  thanks to Cluster-GCN and the proposed normalization method, 
We compare the testing accuracy with other existing methods in Table \ref{tab:stoa}.
For PPI, Cluster-GCN can achieve the state-of-art result by training a 5-layer GCN with 2048 hidden units.
For Reddit, a 4-layer GCN with 128 hidden units is used.

% ppi_deep_4l_vrgcn 0.9869083411336902 816.1084132194519
% ppi_deep_3l_vrgcn 0.9810522490720792 324.67255997657776
% ppi_deep_6l_vrgcn 0.9847758693643515 3352.175052165985
% ppi_deep_2l_vrgcn 0.9088155698591095 125.92578554153442
% ppi_deep_5l_vrgcn 0.9878810246847232 1676.5576736927032

\begin{table}
\caption{Comparisons of running time when using different numbers of GCN layers. We use PPI and run both methods for 200 epochs.}
  \centering
  \renewcommand\bfdefault{b}
  \vspace{-10pt}\begin{tabular}{|l|r|r|r|r|r|}
  \hline
   & 2-layer & 3-layer & 4-layer & 5-layer & 6-layer\\
  \hline
  %Val. F1 & 0.8970  & 0.9750  & 0.9812  & 0.9814  & 0.9802\\
  %\hline
  Cluster-GCN   & 52.9s   & 82.5s   & 109.4s   & 137.8s   & 157.3s\\
  \hline
  VRGCN   & 103.6s   & 229.0s   & 521.2s   & 1054s  & 1956s\\
  \hline
%\news & 19,996 & 1,355,191 & 0.03\% & ? \\
\end{tabular}
\label{tab:deep}
\vspace{-10pt}
\end{table}
\begin{table}
\caption{State-of-the-art performance of testing accuracy reported in recent papers.}
\label{tab:stoa}
  \centering
  \vspace{-10pt}
  \renewcommand\bfdefault{b}
  \begin{tabular}{|l|l|l|}
  \hline
   & PPI & Reddit\\
  \hline
  FastGCN \cite{chen2018fastgcn} & N/A & 93.7\\
  \hline
  GraphSAGE \cite{hamilton2017inductive} & 61.2 & 95.4\\
  \hline
  VR-GCN \cite{chen2018stochastic} & 97.8 & 96.3\\
  \hline
  GaAN \cite{zhang2018gaan}& 98.71 & 96.36\\
  \hline
  GAT \cite{velickovic2018graph} & 97.3 & N/A\\
  \hline
  GeniePath \cite{liu2019geniepath} & 98.5 & N/A\\
  \hline
  Cluster-GCN & \textbf{99.36} & \textbf{96.60} \\
  \hline
% 99.36 -> 5-layer 2048 hidden units
% 99.33 -> 5-layer 1024
\end{tabular}
\vspace{-10pt}
\end{table}

\begin{table*}
\caption{Comparisons of using different diagonal enhancement techniques. For all methods, we present the best validation accuracy achieved in 200 epochs. PPI is used and dropout rate is 0.1 in this experiment. Other settings are the same as in Section~\ref{sec:median}. The numbers marked red indicate poor convergence.}
  \centering
  \renewcommand\bfdefault{b}
  \vspace{-10pt}
  \begin{tabular}{|l|r|r|r|r|r|r|r|}
  \hline
   & 2-layer & 3-layer & 4-layer & 5-layer & 6-layer & 7-layer & 8-layer\\
  \hline
  %Val. F1 & 0.8970  & 0.9750  & 0.9812  & 0.9814  & 0.9802\\
  %\hline
  Cluster-GCN with \eqref{eq:gcn0}  &  90.3  & 97.6 & 98.2  & 98.3 & 94.1 & \textcolor{red}{65.4}  & \textcolor{red}{43.1}\\
  \hline
  Cluster-GCN with \eqref{eq:normalization-id}   & 90.2   & 97.7   & 98.1   & 98.4  & \textcolor{red}{42.4} & \textcolor{red}{42.4} & \textcolor{red}{42.4}\\
  \hline
  Cluster-GCN with \eqref{eq:normalization-id} + \eqref{eq:diag-enhancement}   &  84.9  & 96.0   & 97.1   & 97.6  & 97.3 & \textcolor{red}{43.9} & \textcolor{red}{43.8}\\
  \hline
  Cluster-GCN with \eqref{eq:normalization-id} + \eqref{eq:diag-enhancement-modified}, $\lambda=1$   & 89.6   & 97.5   & 98.2   & 98.3  & 98.0 & 97.4 & 96.2\\
  \hline
\end{tabular}
\label{tab:deep-accuracy}
\end{table*}

% \begin{table}
% \caption{State-of-the-art performance of testing accuracy reported in the papers. {\color{red}(Cho: where did we talk about this table?)}}
% \label{tab:stoa}
%   \centering
%   \renewcommand\bfdefault{b}
%   \begin{tabular}{|c|c|c|c|c|}
%   \hline
%   Methods & GraphSAGE & FastGCN & VR-GCN & Cluster-GCN\\
%   \hline
%   PPI & 61.2  & N/A  & 97.8  & \textbf{99.36}  \\
%   \hline
%   Reddit & 95.4 &  93.7  & 96.3  & \textbf{96.5} \\
% \hline
% % 99.36 -> 5-layer 2048 hidden units
% % 99.33 -> 5-layer 1024
% \end{tabular}
% \end{table}

% \begin{figure}[h]
%     \centering
%     \includegraphics[width=0.4\textwidth]{figures/ppi.png}
%     \includegraphics[width=0.4\textwidth]{figures/reddit.png}
%     \caption{}
% \end{figure}

\section{Conclusion}
We present ClusterGCN, a new GCN training algorithm that is fast and memory efficient. Experimental results show that this method can train very deep GCN on large-scale graph, for instance on a graph with over 2 million nodes, the training time is less than an hour using around 2G memory and achieves accuracy of 90.41 (F1 score).
Using the proposed approach, we are able to successfully train much deeper GCNs, which achieve state-of-the-art test F1 score on PPI and Reddit datasets. 

{\bf Acknowledgement}
CJH acknowledges the support of NSF via IIS-1719097, Intel faculty award, Google Cloud and Nvidia.

\bibliography{cluster_gcn}
\bibliographystyle{ACM-Reference-Format}

\clearpage

\clearpage
%\newpage
\begin{table}
\centering
      \caption{The training, validation, and test splits used in the experiments. Note that for the two amazon datasets we only split into training and test sets.}
      \label{tab:parameter2}
    \begin{tabular}{|l|r|r|r|r|}
    \hline
    Datasets & Task & Data splits (Tr./Val./Te.) \\
     \hline
     PPI  & Inductive & 44906/6514/5524 \\
     \hline
     Reddit  & Inductive & 153932/23699/55334 \\
     \hline
     Amazon & Inductive & 91973/242890 \\
     \hline
  Amazon2M  & Inductive & 1709997/739032 \\
  \hline
    \end{tabular}
  \end{table}

\section{More Details about the experiments}
\label{sec:more-details}
In this section we describe more detailed settings about the experiments to help in reproducibility.

\subsection{Datasets and software versions}

We describe more details about the datasets in Table~\ref{tab:parameter2}.
We download the datasets PPI, Reddit from the website\footnote{\url{http://snap.stanford.edu/graphsage/}} and Amazon from the website\footnote{\url{https://github.com/Hanjun-Dai/steady_state_embedding}}.
Note that for Amazon, we consider GCN in an inductive setting, meaning that the model only learns from training data. In \cite{dai2018learning} they consider a transductive setting.
Regarding software versions, we install CUDA 10.0 and cuDNN 7.0.
TensorFlow 1.12.0 and PyTorch 1.0.0 are used.
We download METIS 5.1.0 via the offcial website\footnote{\url{http://glaros.dtc.umn.edu/gkhome/metis/metis/download}} and use a Python wrapper\footnote{\url{https://metis.readthedocs.io/en/latest/}} for METIS library.

\subsection{Implementation details}
Previous works \citep{chen2018fastgcn, chen2018stochastic} propose to pre-compute the multiplication of $AX$ in the first GCN layer.
We also adopt this strategy in our implementation.
%This strategy indeed gives lots of benefits in terms of accuracy and time.
By precomputing $AX$, we are essentially using the exact 1-hop neighborhood for each node and the expensive neighbors searching in the first layer can be saved.

Another implementation detail is about the technique mentioned in Section~\ref{sec:multiple}
When multiple clusters are selected, some between-cluster links are added back.
Thus the new combined adjacency matrix should be re-normalized to maintain numerical ranges of the resulting embedding matrix.
From experiments we find the renormalization is helpful.

As for the inductive setting, the testing nodes are not visible during the training process.
Thus we construct an adjacency matrix containing only training nodes and another one containing all nodes.
Graph partitioning are applied to the former one and the partitioned adjacency matrix is then re-normalized.
Note that feature normalization is also conducted.
To calculate the memory usage, we consider \texttt{tf.contrib.memory\_stats.BytesInUse()}  for TensorFlow and \texttt{torch.cuda.memory\_allocated()} for PyTorch.

\subsection{The running time of graph clustering algorithm and data preprocessing}

The experiments of comparing different GCN training methods in Section~\ref{sec:exp} consider running time for training.
The preprocessing time for each method is not presented in the tables and figures.
While some of these preprocessing steps such as data loading or parsing are shared across different methods, some steps are algorithm specific.
For instance, our method needs to run graph clustering algorithm during the preprocessing stage.
%Also, since the graph clustering algorithm (METIS) is applied during the data preprocessing procedure, its running time is not shown in Section~\ref{sec:exp}.

In Table~\ref{tab:clustering}, we present more details about preprocessing time of Cluster-GCN on the four GCN datasets. For graph clustering, we adopt Metis, which is a fast and scalable graph clustering library.
We observe that the graph clustering algorithm only takes a small portion of preprocessing time, showing a small extra cost while applying such algorithms and its scalability on large data sets.
In addition, graph clustering only needs to be conducted once to form the node partitions, which can be re-used for later training processes.

\begin{table}
\centering
      \caption{The running time of graph clustering algorithm (METIS) and data preprocessing before the training of GCN.}
      \label{tab:clustering}
    \begin{tabular}{|l|r|r|r|r|}
    \hline
    Datasets & \#Partitions & Clustering & Preprocessing \\
     \hline
     PPI  & 50 & 1.6s & 20.3s \\
     \hline
     Reddit  & 1500 & 33s & 286s \\
     \hline
     Amazon & 200 & 0.3s & 67.5s \\
     \hline
  Amazon2M & 15000 & 148s & 2160s \\
  \hline
    \end{tabular}
  \end{table}

\end{document}